\documentclass{article}
\usepackage{ijcai26}

\usepackage{times}
\usepackage{soul}
\usepackage{url}
\usepackage[hidelinks]{hyperref}
\usepackage[utf8]{inputenc}
\usepackage[small]{caption}
\usepackage{graphicx}
\usepackage{amsmath}
\usepackage{amsthm}
\usepackage{booktabs}
\usepackage{algorithm}
\usepackage{algorithmic}
\usepackage[switch]{lineno}
\usepackage{enumitem}
\usepackage{listings}
\lstdefinelanguage{json}{
  basicstyle=\ttfamily,
  showstringspaces=false,
  breaklines=true,
  breakatwhitespace=false,
  columns=fullflexible,
  keepspaces=true,
  morestring=[b]",
  stringstyle=\ttfamily,
  literate=
   *{0}{{0}}1 {1}{{1}}1 {2}{{2}}1 {3}{{3}}1 {4}{{4}}1
    {5}{{5}}1 {6}{{6}}1 {7}{{7}}1 {8}{{8}}1 {9}{{9}}1
    {:}{{:}}1 {,}{{,}}1 {\{}{{\{}}1 {\}}{{\}}}1 {[}{{[}}1 {]}{{]}}1
}
\usepackage{xcolor}
\usepackage{wrapfig}
\usepackage{multirow}

\newcommand{\method}[0]{\text{L4L{\ }}}

\title{Towards Trustworthy Legal AI through LLM Agents and Formal Reasoning}

\author{
Chen Linze$^1$
\and
Cai Yufan$^{1}$\and
Hou Zhe$^{2}$\And
Dong Jin Song$^1$\\
\affiliations
$^1$National University of Singapore\\
$^2$Griffith University\\
\emails
e1344652@u.nus.edu,
caiyf@nus.edu.sg,
z.hou@griffith.edu.au,
dcsdjs@nus.edu.sg
}

\begin{document}

\maketitle

\begin{abstract}
Legal decisions should be logical and based on statutory laws. While large language models (LLMs) are good at understanding legal text, they cannot provide verifiable justifications.
We present \method, a solver-centric framework that enforces formal alignment between LLM-based legal reasoning and statutory laws.
The framework integrates role-differentiated LLM agents with SMT-backed verification, combining the 
flexibility of natural language with the rigor of symbolic reasoning.
Our approach operates in four stages:
(1) Statute Knowledge Building, where LLMs autoformalize legal provisions into logical constraints and validate them through case-level testing;
(2) Dual Fact-and-Statute Extraction, in which prosecutor- and defense-aligned agents independently map case narratives to argument tuples;
(3) Solver-Centric Adjudication, where SMT solvers check the legal admissibility and consistency of the arguments against the formalized statute knowledge;
(4) Judicial Rendering, in which a judge agent integrates solver-validated reasoning with statutory interpretation and 
similar
precedents to produce a legally grounded verdict.
Experiments on public legal benchmarks show that \method consistently outperforms baselines, while providing auditable justifications that enable trustworthy legal AI.
\end{abstract}

\section{Introduction}
Legal decision-making requires more than accurate linguistic interpretation of statutory text.
In rule-of-law systems, judicial outcomes are expected to satisfy \emph{formal rationality}~\cite{linna2025judicial}.
Conclusions must be justified by demonstrating their consistency with explicitly stated,
general, and logically coherent legal rules~\cite{sadowski2025verifiable}.
Courts are therefore constrained not to select outcomes based on intuitive or moral preference alone,
but to articulate decisions that can be traced back to governing statutes, interpretations, and precedents~\cite{kesari2024legal}.

\smallskip
\noindent\textbf{Current Limitations.}
Recent LLMs have shown impressive performance on legal tasks,
including opinion summarization, pleading drafting~\cite{hendrycks2021measuring},
and bar-exam question answering~\cite{katz2023gpt}.
However, purely neural approaches are prone to hallucinating authorities~\cite{chen2023hallucination},
conflating distinct doctrinal requirements~\cite{savelka2023when},
and producing conclusions whose logical validity cannot be independently verified.
Domain-adapted systems such as \textsc{ChatLaw}~\cite{cui2023chatlaw},
\textsc{LawLLM}~\cite{zheng2024lawllm}, and \textsc{Lexi}~\cite{lehmberg2025lexi}
reduce factual errors through retrieval-augmented generation,
yet they ultimately rely on opaque generation processes and do not provide guarantees.

\smallskip
\noindent\textbf{Legal Formalization.}
The use of formal representations for legal norms has long been a central research direction
in Law~\cite{richmond2024explainable}.
Importantly, legal formalization does not claim to exhaust the full semantic or moral content of law.
Rather, it provides a \emph{controlled abstraction} of explicitly articulated statutes and rules
that can be subjected to logical scrutiny~\cite{prajescu2025argumentation}.
At the same time, legal decision-making inevitably involves judicial discretion,
particularly in the presence of vague concepts, competing principles, or unforeseen factual patterns.
Such discretion does not imply the absence of structure,
but reflects the existence of alternative interpretations and
subsequent amendments that refine base statutes~\cite{foy2010judicial}.
A trustworthy legal AI system needs therefore not eliminate discretion,
but make interpretive choices explicit and auditable.

\smallskip
\noindent\textbf{Bridging Substantive and Formal Rationality.}
In this work, we present \textbf{\method},
a solver-centered framework for bridging substantive and formal rationality
in LLM-based legal reasoning.
\method embeds role-differentiated LLM agents within a formal reasoning loop
governed by formalized statutory constraints.
Legal norms are represented using a typed schema
that captures actors, actions, conditions, and deontic status,
and are compiled into executable SMT constraints.
LLM agents aligned with prosecutorial and defense roles
independently extract case facts and candidate statutes.
An SMT solver then adjudicates the resulting constraint system,
verifying logical consistency and identifying misalignment
between competing arguments and statutory requirements.

Crucially, \method separates adjudication from rendering.
Formal reasoning outcomes produced by the solver are not directly exposed to end users.
Instead, a judge agent performs \emph{judicial rendering} by integrating solver-validated results
with statutory interpretation principles and retrieved analogous precedents,
producing a transparent and legally grounded verdict.
This design preserves the interpretive flexibility required for substantive rationality,
while ensuring that final conclusions remain anchored in formally verified legal constraints.

We evaluate \method on public legal benchmarks~\cite{li2023lecardv2}~\cite{xue2024leec},
demonstrating strong improvements over LLM baselines,
in statute selection, verdict accuracy, and sentencing quality.
Crucially, our system produces solver-checked symbolic justifications
that enable auditability of legal conclusions.


\paragraph{Contributions.}
\begin{itemize}
    \item 
    We propose a systematic approach to formalizing statutory law, in which natural-language legal provisions are translated into executable logical constraints with explicit semantic structure.

    \item 
    We introduce a SMT solver-centered legal reasoning framework that bridges substantive and formal rationality by integrating formal reasoning into the adjudication.

    \item 
    We design a role-differentiated legal agent architecture, in which prosecutor- and defense-aligned agents independently extract facts and argue statute applicability under shared formal constraints.

    \item
    We conduct extensive experiments on multiple benchmarks, demonstrating that our approach achieves higher performance and robustness than other legal AI systems.
\end{itemize}
\section{Approach}
\label{sec:approach}
\begin{figure*}
    \centering
    \includegraphics[width=1.0\linewidth]{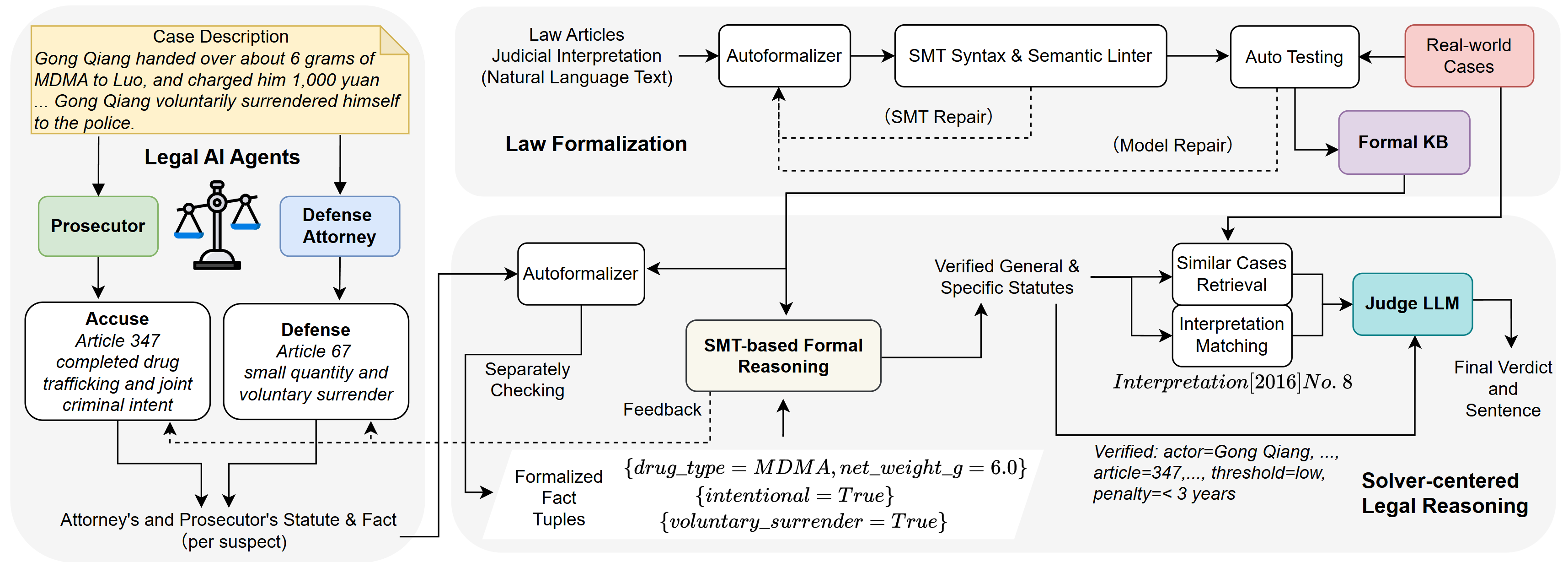}
    \caption{Overview of \method, a solver-centered framework for auditable legal
reasoning.
(\textbf{Top}) \emph{Law formalization}: natural-language text are automatically translated into SMT constraints, with comprehensive checking and testing, to construct a verified formal knowledge base.
(\textbf{Left}) \emph{Legal AI Agents}: given a case description, prosecutor
and defense LLM agents independently extract suspect-centric facts and propose
applicable statutes.
(\textbf{Bottom}) \emph{Legal reasoning}: 
Agent outputs are autoformalized into typed fact tuples and checked separately against the formal KB.
An SMT solver verifies article-level applicability and then determines
the satisfied statutory clauses, producing legally admissible conclusions.
Verified statutes, with similar cases and interpretation,
are finally rendered by a judge LLM into the final verdict and sentence.}
    \label{fig:framework}
\end{figure*}
Our framework couples the abilities of LLMs with the proof obligations enforced by an SMT solver (Z3) in order to deliver auditable legal conclusions.  

\subsection{Law Formalization}
\label{subsec:formalization}

\paragraph{Meta-schema.}
We start from a universal first-order template that captures the quadruple \emph{Actor–Action–Condition–Norm}, with penalty attached:
\begin{align}
& \forall a{:}{\sf Actor},\,\alpha{:}{\sf Action},\,\kappa{:}{\sf Condition},\,n{:}{\sf Norm},\,
      p{:}{\sf Penalty}. \nonumber\\
& {\sf scope}(a,\alpha,\kappa,n)\;\rightarrow\;
{\sf applies}(a,\alpha,\kappa,n,p).
\end{align}

The sort \({\sf Norm}\) is an element of \texttt{prohibited}, \texttt{obligated}, \texttt{permitted}, while
\({\sf scope}\) will be progressively refined in the subsequent layers.
This schema paritally follows the previous work \cite{xue2024leec}.
The schema is globally consistent across all statutes, and every constraint field in downstream verification has a corresponding default extraction output. 

\paragraph{Article instantiation.}
For a concrete statute article \(\text{Art}_i\) we introduce an \emph{article guard}~\({\sf art\_i}\) that fixes the overall applicability domain (jurisdiction, offense category, subject status,~\emph{etc.}). 
The formalized article is connected with the meta-schema by
${\sf scope}\!\bigl(
  \langle{\sf prohibited},\text{Art}_i\rangle,
  a,\alpha,\kappa
\bigr)
\;:\equiv\;
{\sf art\_i}(a,\alpha,\kappa).$

\paragraph{Clause instantiation.}
Each article typically contains several clauses that refine quantitative
thresholds or aggravating circumstances.
Let \(\text{Art}_i.c_j\) denote clause~\(j\) of article~\(i\).
For each article, we define a \emph{clause guard}~\({\sf cl_{i,j}}\) together with a statutory
penalty range~\(p_{i,j}\). 

The resulting Horn clauses are
\begin{align}
&\forall a,\alpha,\kappa. \; {\sf art\_i}(a,\alpha,\kappa)\;\land\;
  {\sf cl_{i,j}}(\kappa) \nonumber\\
&\rightarrow\;
  {\sf prohibited}(a,\alpha,\kappa)\;\land\;
  {\sf penalty}(a,\alpha,\kappa)=p_{i,j}.
\end{align}

\textbf{Formal Knowledge Base (KB).}
Each article in the criminal code is compiled into the above form, with cross-references (e.g., ``Article~347(1)--(3)'') explicitly disambiguated and key quantities such as drug amounts and monetary thresholds parameterized.
The construction of the statute and judicial interpretation KB follows a
neurosymbolic formalization-and-testing paradigm.
As illustrated in Figure~\ref{fig:framework}, natural-language statutes and judicial interpretations are automatically formalized into SMT models.
These SMT models are subsequently validated through syntactic checks and semantic linting to ensure well-formedness and logical soundness.
We further evaluate the correctness of the formalized models using real-world legal cases with known outcomes, and iteratively repair the SMT constraints when solver results largely deviate from the corresponding gold labels.
This process grounds the induced statutory constraints in both formal validity and empirical legal practice.

\subsection{Legal AI Agents}
\label{subsec:legal-llm}
Figure~\ref{fig:framework} (left half) depicts two role-differentiated LLM agents, prosecutor and defense attorney, that mirror the real courtroom.  
The pipeline is in the following steps:

\paragraph{Suspect-Centric Decomposition.}
Criminal cases often involve multiple suspects whose roles, intent, and
liability may differ substantially.
Given a case narrative, \method first decomposes it into a set of
suspect-specific views, and all subsequent fact extraction and statute
reasoning are performed independently for each suspect.
This design ensures that evidential facts, statute applicability, and
sentencing decisions are not conflated across different defendants.

\paragraph{Dual Fact–\&–Statute Extractors.}
For each suspect, the \textsc{Prosecutor LLM} and \textsc{Attorney LLM} receive the same free-text case narrative but are prompted with adversarial instructions (``maximise conviction’’ vs.\ ``maximise acquittal’’).  
The dual-extractor design prevents a single model from blending prosecution and defense biases.
Each produces:
\begin{itemize}[leftmargin=1.2em, itemsep=0pt]
  \item Fact tuples
        \(\mathcal{F}\langle a,\alpha,\kappa,w\rangle\), which instantiate the Actor--Action--Condition components of the shared legal schema, where \(w\in[0,1]\) encodes evidential confidence;
  \item Candidate statute ranks
        \(S=[(id_1,\pi_1),\dots]\) whose prior probabilities
        \(\pi_i\) stem from retrieval over the formal KB using dense embeddings.
\end{itemize}
\subsection{Solver-Centered Legal Reasoning}
\label{sec:solver_reasoning}

\method is organized around a \emph{solver-centered} legal reasoning pipeline,
in which symbolic satisfiability and optimization serve as the authoritative
mechanisms for validating legal applicability and deriving legally admissible
outcomes.
All legal conclusions are produced only if they satisfy the formal statutory
constraints encoded in the SMT solver.

\paragraph{Autoformalizer LLM.}
A role-neutral LLM serves as a schema-guided interface between the dual fact
extractors and the formal reasoning backend.
For each suspect, it transforms the structured outputs
$\langle\mathcal{F}, S\rangle$ into a well-typed SMT constraint set $\Phi$ by
instantiating the shared meta-schema and querying the formal knowledge base
$\mathcal{K}$.
Concretely, the autoformalizer performs the following steps:
\begin{itemize}[leftmargin=1.2em, itemsep=0pt]
    \item \textbf{Value grounding}, converting linguistic or fuzzy quantities
    (e.g., ``large amount'', ``multiple transactions'') into symbolic variables
    and numeric constraints;
    \item \textbf{Statute resolution}, expanding statutory references and clause
    combinations (e.g., ``Art.~347(1)\&(3)'') into guarded Horn clauses defined
    in the formal KB;
    \item \textbf{Provenance tagging}, attaching agent identifiers and confidence
    weights to each asserted fact and statute, enabling traceability and
    conflict diagnosis.
\end{itemize}

The resulting constraint system is
\[
\Phi \;=\; \Phi_{\text{facts}} \;\cup\;
           \Phi_{\text{statutes}} \;\cup\;
           \Phi_{\text{guards}},
\]
which constitutes the sole input to the solver.

\paragraph{Formal Legal Reasoning.}
Formal legal reasoning is conducted in two successive solver stages,
corresponding to statutory applicability and clause-level qualification.

\paragraph{Stage I: Article applicability verification.}
We first verify whether a candidate statutory article is applicable to the
extracted facts.
Formally, for each article $\text{Art}_i$, we check the satisfiability of
$\Phi \;\land\;
\exists a,\alpha,\kappa.\;
{\sf art\_i}(a,\alpha,\kappa),$
where ${\sf art\_i}$ denotes the article-level guard encoding the scope of
$\text{Art}_i$.
If the solver returns \textsf{unsat}, the article is deemed inapplicable and
discarded from further consideration.

\paragraph{Stage II: Clause qualification.}
For each applicable article, we further determine which clause(s)
$\text{Art}_i.c_j$ are satisfied by the facts.
For each clause guard ${\sf cl}_{i,j}$, check
\[
\Phi \;\land\;
{\sf art\_i}(a,\alpha,\kappa)
\;\land\;
{\sf cl}_{i,j}(\kappa).
\]
A clause $\text{Art}_i.c_j$ is selected if the above formula is satisfiable.
Each selected clause uniquely determines the corresponding legal consequence,
including offense qualification and statutory penalty range $p_{i,j}$.
When multiple clauses are simultaneously satisfiable, 
the solver enumerates all admissible clauses, 
leaving discretionary selection to downstream judicial policies.

\paragraph{Iterative feedback loop.}
If either applicability verification or judgment derivation fails
(i.e., the solver reports \textsf{unsat} or \textsf{unknown}),
the pipeline does not produce a verdict.
Instead, the minimal unsat core is returned to the autoformalizer, which
identifies whether the inconsistency arises from fact extraction errors,
statute over-approximation, or incompatible clause combinations.
The corresponding constraints are revised, and solver-centered reasoning is
re-invoked until a legally consistent outcome is obtained or all candidate statutes are exhausted.

\subsection{Judge LLM and Rendering Mechanism.}
Once SMT-solver returns a \textsf{sat} model, its proof object together
with the surviving statutes and sentencing ranges are passed to the
\textsc{Judge LLM}.
The Judge LLM drafts 
(i) a declarative verdict (\textit{guilty} / \textit{not guilty}) and the statutes, 
(ii) a quantified sentence derived from the SMT solver’s ranges, and 
(iii) a narrative justification that references solver-derived facts, the relevant Code articles, judicial interpretations, and each party’s key arguments.

\paragraph{Similar Case Retrieval.}
In parallel with formal reasoning, \method retrieves similar cases
based on the verified statutes and case descriptions.
Similar cases are not used to override solver results, but are injected as auxiliary context to the Judge LLM.
This design reflects real-world judicial practice, 
where precedent and analogous cases inform the exercise of judicial discretion without altering statutory applicability.

\paragraph{Judicial Interpretation Matching.}
Judicial interpretation is conditioned on the set of verified statutes and serves to refine their legal meaning in specific contexts.
It is injected into the prompt of Judge LLM using the statute matching.
This separation allows the core reasoning pipeline to remain statute-centric, while enabling flexible adaptation to different legal interpretive regimes and evolving jurisprudence.

\paragraph{Summary.}
Our design explicitly combines formal rationality with substantive judgment. 
Formal rationality is enforced by the SMT solver through applicability verification. 
Substantive rationality, including judicial discretion and contextual reasoning, is modelled at the LLM layer through the incorporation of similar cases and judicial interpretations. 
\section{Running Example}
\label{sec:mot-ex}
In this section, we introduce a running example to illustrate our pipeline end-to-end.

\begin{table*}[ht]
\centering
\renewcommand{\arraystretch}{1.1}
\small
\resizebox{\linewidth}{!}{
\begin{tabular}{p{1.6cm} p{6.6cm} p{6.8cm}}
\toprule
\textbf{Article} & \textbf{Formal Schema} & \textbf{Description and Legal Implication} \\
\midrule
347 (Drug Trafficking) &
\textit{Norm} (prohibited);
\textit{Actor} (person, unit,  
Age (integer, $\geq$12 years, default $\geq$18), ...);
\textit{Action} (smuggling, selling, transporting, manufacturing);  
\textit{Condition} (
DrugQuantity (opium/heroin/meth thresholds in grams);  
Circumstance (ringleader, armed protection, international trafficking);  
);
\textit{Penalty}(...) &
Defines factors affecting sentencing for narcotics crimes, including actor type, age, type of drug activity, drug quantity thresholds, and aggravating circumstances (e.g., organized or armed operations). These parameters determine the severity level of Article 347 sentencing rules. \\
\bottomrule
\end{tabular}}
\caption{Examples of Law-Specific Knowledge Base.}
\label{tab:article_kb_r}
\end{table*}

\paragraph{Case Synopsis.}
\begin{quote}
Defendant Gong Qiang received a WeChat message from Luo, requesting the purchase of 1{,}000 yuan worth of drugs.
After receiving the 1{,}000 yuan transfer via WeChat, Gong Qiang, together with accomplice Zhang, handed over
\textit{10 grams of MDMA} to Luo. 
On Nov 7, 2019, Gong Qiang voluntarily surrendered to the police.
\end{quote}

\paragraph{Formal Statute and Judicial Interpretation.}
As shown in \autoref{tab:article_kb_r}, our formal KB contains both \emph{criminal-law articles} (e.g., Article~347 on trafficking; Article~65 on recidivism; Article~67 on surrender/confession)
and \emph{judicial interpretations} that refine quantitative thresholds.
In particular, the Supreme People's Court interpretation \textbf{Interpretation [2016] No.~8} provides: 
Amphetamine-type drugs such as MDMA, morphine-equivalent more than 20 grams but less than 100 grams $\Rightarrow$ ``\textit{other drugs in relatively large quantity}''
for purposes of Criminal Law Article~347(2)(1) and Article~348.
This interpretation becomes a \emph{quantity-threshold constraint} used to classify the case into the appropriate sentencing band.

\autoref{tab:z3_reasoner_r} illustrates the canonical reasoning template 
to operationalize statute-specific sentencing logic.
For each applicable criminal law article, \method automatically compiles
the corresponding statutory text and judicial interpretations.
This formulation is uniform across statutes and does not rely on
article-specific heuristics.

\paragraph{Role-Differentiated Legal AI Agents.}
We instantiate two aligned agents:
\textit{(i) Prosecutor agent} that tends to maximize statutory coverage and aggravating factors,
and \textit{(ii) Defense agent} that prioritizes exclusion/mitigation and evidentiary uncertainty.
Both agents must express outputs in a shared formal schema.
Given the narrative, the two agents independently extract structured facts and the autoformalizer translated them into SMT assertions.

\noindent(A) Prosecutor (abbreviated):
\begin{itemize}\setlength{\itemsep}{0pt}\setlength{\parskip}{0pt}
    \item Drug facts: \{drug\_type=MDMA, net\_weight\_g=6.0\}
    \item Transaction facts: \{payment\_channel=WeChat, payment\_amount\_yuan=1000, completed=True\}
    \item Participation: \{Gong\_role=organizer\}
    \item Mens rea: \{intentional=True\}
    \item Procedural: \{voluntary\_surrender=True\}
\end{itemize}

\noindent(B) Defense Attorney (abbreviated):
\begin{itemize}\setlength{\itemsep}{0pt}\setlength{\parskip}{0pt}
    \item Quantity band emphasis: \{drug\_type=MDMA, net\_weight\_g=6.0, threshold=low\}
    \item Role emphasis: \{Gong\_role=intermediary\}
    \item Mitigation: \{voluntary\_surrender=True\}
    \item Uncertainty flags: \{purity=unknown\}
\end{itemize}

\paragraph{Formal Legal Reasoning}
Each statute/interpretation is compiled into solver constraints, which are proven satisfiable by an SMT solver and can be exported as verified justification. Below we show the key constraints.
\begin{itemize}\setlength{\itemsep}{0pt}\setlength{\parskip}{0pt}
    \item Art.~347 (Trafficking): \textit{satisfiable} $\Rightarrow$ \textsf{retained}.
    \item Interpretation [2016] No.~8 : $6.0 > 20$ $\Rightarrow$ \textsf{false}.
    \item Art.~67 (Mitigation): if surrender/confession facts are consistent $\Rightarrow$ \textsf{retained}.
    \item Art.~65 (Recidivism): \textsf{retained} \emph{only if} prior-sentence and 5-year window are supported by admissible evidence;
    otherwise the slice becomes \textsf{unknown}.
\end{itemize}

\begin{table*}[ht]
\centering
\renewcommand{\arraystretch}{1.15}
\small
\begin{tabular}{p{0.48\linewidth} p{0.48\linewidth}}
\toprule
\textbf{Algorithm} & \textbf{Example Execution: Article 347 (Drug Trafficking)} \\
\midrule
\textbf{Input:} Statute-specific fact slice $f$, compiled rule set $\Phi$ \newline
\textbf{Output:} Applicable sentencing clause $C$ \newline
1: Initialize constraint solver $S$; set $C=\emptyset$ \newline
2: Declare legal entities (actors, actions, ...) \newline
3: Encode factual predicates from $f$ into $S$ \newline
4: Sort rules $r_i \in \Phi$ by descending statutory severity \newline
5: \textbf{for} each $r_i$ \textbf{do} \newline
6:\quad Push solver state \newline
7:\quad Encode statutory and interpretative constraints of $r_i$ \newline
8:\quad Check satisfiability of $r_i$ under $S$ \newline
9:\quad \textbf{if} SAT and $C=\emptyset$ \textbf{then} \newline
10:\quad\quad Assign $C \leftarrow$ consequence of $r_i$ \newline
11:\quad\quad Pop solver state and \textbf{break} \newline
12:\quad Pop solver state \newline
13: \textbf{if} $C=\emptyset$ \textbf{then} $C \leftarrow$ ``No applicable clause'' \newline
14: \textbf{return} $C$
&
1: Identify Actor = Guo Qiang; Action = trafficking drugs \newline
2: Encode facts (DrugType = MDMA, NetWeight = 6g, Intentional = True) \newline
3: Iterate over Article 347 clauses in descending severity \newline
4: Push state for $r_1$ (large quantity) \newline
5: Check $r_1$ $\Rightarrow$ UNSAT (quantity threshold not met) \newline
6: Push state for $r_2$ (medium quantity) \newline
7: Check $r_2$ $\Rightarrow$ UNSAT (quantity threshold not met) \newline
8: Push state for $r_3$ (small quantity) \newline
9: Check $r_3$ $\Rightarrow$ SAT \newline
10: Push state for $r_4$ (small quantity + different circumstance) \newline
...\newline
14: Assign $C =$ ``Fixed-term imprisonment $\leq$ 3 years'' \newline
15: Discard other solver states \newline
16: Verify non-empty C, ready for return \newline
17: Return final sentencing clause under Article 347 \\
\bottomrule
\end{tabular}
\caption{Canonical Solver-Based Reasoning Template for Statutory Sentencing
(Algorithm and Example Execution for Article 347).}
\label{tab:z3_reasoner_r}
\end{table*}

\paragraph{Final Judge Rendering.}
By explicitly matching Fa Shi [2016] No.~8, the system classify MDMA as \textit{other drugs} in the Article 347 and enforces the quantitative guard.
This prevents miss-charging or miss-sentencing caused by other undefined drugs and the thresholds.
After the solver fixes the applicable charge and quantity band, the system can retrieve similar precedents with the same drug type (MDMA), comparable weight range (e.g., 1--10g), and similar procedural posture, to produce stable sentencing rationales.
Finally, the judge agent generates an auditable draft.
In the example, a prompt-only judge tended to overestimate (``2 years''),
while \method moves the recommendation toward the ground-truth range (``around 1 year'').
\section{Experiments}
Specifically, we investigate the following research questions:
\begin{itemize}
    \item \textbf{RQ1}: How does our method perform compared to state-of-the-art baselines?
    \item \textbf{RQ2}: What is the contribution of different components of our method to final sentencing accuracy?
    \item \textbf{RQ3}: Does our method provide better robustness?
\end{itemize}

\paragraph{Datasets.}
We conduct our experiments on three datasets covering real-world cases
and controlled perturbations. 
Detailed dataset preprocessing, and annotation procedures are provided in the appendix among supplementary material. 
Table~\ref{tab:dataset_stats} shows the dataset statistics, including case counts, average lengths, and the average number of statutes.
\begin{itemize}
    \item \textbf{LeCaRDv2 Dataset}~\cite{li2023lecardv2}, a large-scale
    Chinese criminal case dataset with statutes and judgments;

    \item \textbf{LEEC Dataset}~\cite{xue2024leec}, a legal element extraction
    corpus derived from publicly available Chinese criminal judgments,
    which we further process to support suspect-level evaluation;

    \item \textbf{Synthetic Perturbation Dataset}, constructed by injecting
    legally meaningful factual modifications to assess robustness under
    controlled statute-triggering changes.
\end{itemize}

\begin{table}[ht]
\centering
\renewcommand{\arraystretch}{1.1}
\small
\begin{tabular}{lccc}
\toprule
Dataset & \#Cases & Avg. Length & Avg. \#Statutes \\
\midrule
LeCaRDv2 & 55192 & 889.17 & 4.53 \\
LEEC (Suspect-Level)  & 9470 & 654.49 & 5.25 \\
Perturbed Dataset & 3622 & 77.36 & 4.88 \\
\bottomrule
\end{tabular}
\caption{Statistics of Experimental Datasets.}
\label{tab:dataset_stats}
\end{table}


\paragraph{Baselines.}
The baselines cover both pure LLM inference, specialized legal agents and retrieval-augmented
settings to ensure a comprehensive comparison, including GPT-4o, GPT o4-mini, GPT-5.2, DeepSeek V3, Claude 4 Sonnet, DISC-Law~\cite{yue2023disc} and LexiLaw~\cite{LexiLaw}.

\paragraph{Settings.}
We adopt GPT-5.2 throughout the entire pipeline of \method.
Our experiments evaluate two aspects of performance:
(i) accuracy on real-world legal cases, and
(ii) robustness under controlled factual perturbations.
Across all datasets, we construct disjoint subsets for
(i) formal knowledge base validation,
(ii) retrieval and case matching, and
(iii) final evaluation.
Unless otherwise specified, each subset is obtained by random sampling
under fixed seeds to ensure fair and reproducible comparison.
Evaluation on LeCaRDv2 is performed at the case level, while LEEC is evaluated at the suspect level, with an additional metric for multi-suspect identification.
For robustness evaluation, we use the Synthetic Perturbation Dataset and measure the change Accuracy, which quantifies whether the model correctly captures changes induced by designed factual perturbations.
Implementation details and the code are provided in the supplementary material.

\begin{table*}[ht]
\centering
\renewcommand{\arraystretch}{1.1}
\small
\begin{tabular}{lcccccccccccc}
\toprule
 & \multicolumn{6}{c}{LeCaRDv2} & \multicolumn{6}{c}{LEEC} \\
\cmidrule(lr){2-7} \cmidrule(lr){8-13}
 & \multicolumn{3}{c}{General} & \multicolumn{3}{c}{Specific}
 & \multicolumn{3}{c}{General} & \multicolumn{3}{c}{Specific} \\
\cmidrule(lr){2-4} \cmidrule(lr){5-7}
\cmidrule(lr){8-10} \cmidrule(lr){11-13}
Model
& P & R & F1 & P & R & F1
& P & R & F1 & P & R & F1 \\
\midrule
LexiLaw
& 8.96 & 26.27 & 13.36 & 47.76 & 49.91 & 48.81
& 1.03 & 0.21 & 0.34 & 1.89 & 3.09 & 2.23 \\

DISC-LawLLM
& \textbf{66.67} & 2.13 & 4.12 & 64.22 & 75.27 & 69.31
& 0.00 & 0.00 & 0.00 & 1.56 & 1.25 & 1.04 \\

GPT o4-mini
& 21.54 & 35.00 & 26.67 & 69.00 & 21.00 & 32.00
& 34.80 & 23.24 & 25.52 & 65.93 & 63.37 & 62.82 \\

GPT-4o
& 12.00 & 36.00 & 18.00 & 67.00 & 73.00 & 70.00
& 34.00 & 23.27 & 24.85 & 67.15 & 66.58 & 65.22 \\

Claude 4 Sonnet
& 32.79 & 26.67 & 29.41 & 64.00 & 75.00 & 69.00
& 36.85 & 24.05 & 26.05 & 65.96 & 63.63 & 63.73 \\

DeepSeek v3
& 10.25 & 44.84 & 16.88 & 60.19 & \textbf{82.77} & 69.70
& 32.16 & 22.95 & 23.29 & 64.72 & 63.49 & 62.50 \\

GPT-5.2
& 23.47 & \textbf{46.25} & 31.14 & 70.71 & 81.05 & 75.53
& 52.01 & 37.38 & 39.08 & 79.73 & 76.06 & 76.46 \\


\textbf{\method~(Ours)}
& 34.00 & 45.95 & \textbf{39.08}
& \textbf{81.03} & 77.05 & \textbf{78.99}
& \textbf{64.05} & \textbf{40.77} & \textbf{45.51}
& \textbf{82.35} & \textbf{76.71} & \textbf{78.13} \\
\bottomrule
\end{tabular}
\caption{Provision Prediction Performance on dataset LeCaRDv2 and LEEC (\%), comparing (P)recision, (R)ecal, and F1 scores.}
\label{tab:rq1_all}
\end{table*}

\subsection{RQ1: Accuracy Evaluation}

We evaluate all models on statute identification and sentencing prediction. 
Statute prediction is formulated as a multi-label classification task, and we report Precision, Recall, and F1-score for both general and specific provisions.
For sentencing prediction, we report Average Sentencing Error (ASE) in Months and Root Mean Squared Error (RMSE) in Months.
We measure a Valid Ratio to assess legal consistency of the predicted outcomes and the formal laws. 
For \method, it measures whether the Judge LLM follows the provided formal reasoning results.
On the LEEC dataset, we report Suspect Extraction F1 to evaluate the accuracy of suspect-level decomposition in multi-defendant cases.
The detailed metrics are in the appendix.
Table~\ref{tab:rq1_all} summarizes provision prediction performance.
On LeCaRDv2, general-purpose LLMs tend to favor recall at the expense of precision, leading to systematic over-prediction of applicable statutes and more recent LLMs further improve recall.
In contrast, \method applies SMT-based verification to filter legally inconsistent arguments, which may reject some marginally applicable provisions but substantially improves precision.
\method maintains a better precision--recall balance and achieves the best F1 score on both general and specific provisions.
The advantage of \method is more pronounced on LEEC dataset.
\method delivers the best precision, recall, and F1 scores in multi-defendant settings, where complex suspect-level cases increases difficulties.
Table~\ref{tab:sentencing_all} summarizes sentencing accuracy, legal validity, and suspect extraction performance.
\method achieves the best overall performance, with the lowest sentencing error, highest Valid Ratio under statute grounding, and the highest Suspect Extraction F1, demonstrating robust suspect-centric reasoning in multi-defendant cases. 
These results underscore the importance of solver-anchored verification for accurate and legally admissible sentencing.

\begin{table*}[ht]
\centering
\renewcommand{\arraystretch}{1.1}
\small
\begin{tabular}{lccccccccc}
\toprule
\multirow{3}{*}{Model}
& \multicolumn{4}{c}{LeCaRDv2}
& \multicolumn{5}{c}{LEEC} \\
\cmidrule(lr){2-5} \cmidrule(lr){6-10}

& \multicolumn{2}{c}{w/o Golden Statute}
& \multicolumn{2}{c}{w/ Golden Statute}
& \multicolumn{3}{c}{w/o Golden Statute}
& \multicolumn{2}{c}{w/ Golden Statute} \\
\cmidrule(lr){2-3} \cmidrule(lr){4-5}
\cmidrule(lr){6-8} \cmidrule(lr){9-10}

& RMSE$\downarrow$ & Valid (\%)$\uparrow$
& RMSE$\downarrow$ & Valid (\%)$\uparrow$
& RMSE$\downarrow$ & Valid (\%)$\uparrow$ & SusF1 (\%)$\uparrow$
& RMSE$\downarrow$ & Valid (\%)$\uparrow$ \\
\midrule

LexiLaw
& 39.09 & 92.13 & 28.51 & 90.00
& 25.89 & 97.00 & 9.62
& 31.75 & 96.00 \\

DISC-LawLLM
& 52.45 & 83.00 & 37.14 & 90.00
& -- & -- & 13.87
& -- & -- \\

GPT o4-mini
& 33.20 & 92.00 & 28.03 & 94.00
& 34.94 & 91.00 & 78.68
& 38.46 & 96.00 \\

GPT-4o
& 32.84 & 94.00 & 22.68 & 95.00
& 41.66 & 95.60 & 80.25
& 50.41 & 95.50 \\

Claude 4 Sonnet
& 26.25 & 93.00 & 26.20 & 94.00
& 36.92 & 94.00 & 89.97
& 28.40 & \textbf{98.00} \\

DeepSeek v3
& 26.73 & 64.30 & 15.42 & 93.00
& 27.67 & 97.00 & 66.48
& 22.61 & 97.00 \\

GPT 5.2
& 14.54 & 94.20 & 13.87 & 94.00
& 28.48 & 95.20 & 97.80
& 23.90 & 95.00 \\


\textbf{\method~(Ours)}
& \textbf{12.72} & \textbf{94.60}
& \textbf{9.98} & \textbf{96.20}
& \textbf{23.04} & \textbf{97.20} & \textbf{98.80}
& \textbf{20.95} & 96.77 \\

\bottomrule
\end{tabular}
\caption{Sentencing Error, Legal Validity, and Suspect-Level Performance with or without Golden Statutes}
\label{tab:sentencing_all}
\end{table*}

\subsection{RQ2: Ablation Study}

To quantify the contribution of each component in our Legal LLM Agent framework (Figure~\ref{fig:framework}), 
we conduct an ablation study on LeCaRDv2 by selectively disabling key modules and comparing them against
the full pipeline.
Table~\ref{tab:ablation_results} reports three evaluation metrics:
\textbf{G-F1} for general-provision prediction,
\textbf{S-F1} for specific-provision prediction, and
\textbf{SE (M)} for sentencing error measured in months.
Disabling the \textbf{attorney module} reduces general-provision accuracy and increases sentencing error, indicating that dual-agent adversarial extraction provides complementary perspectives beyond a prosecution-only view.
Disabling the \textbf{formal reasoning module} causes the most severe degradation, with sharp drops in G-F1 and S-F1 and sentencing error, highlighting the necessity of formal logical verification.
Disabling the \textbf{judge rendering module} increases sentencing error, suggesting its primary role in refining sentencing severity.

\begin{table}[ht]
\centering
\renewcommand{\arraystretch}{1.05}
\small
\begin{tabular}{lccc}
\toprule
\textbf{Model Variant} & \textbf{G-F1 (\%)} & \textbf{S-F1 (\%)} & \textbf{SE (M)} \\
\midrule
\textbf{-- Attorney}        & 37.89 & 69.56 & 13.00 \\
\textbf{-- Formal Reasoning}     & 25.29 & 61.01 & 20.87 \\
\textbf{-- Judge Rendering}   & \textbf{39.08} & \textbf{78.99} & 11.88 \\
Raw GPT-5.2                 & 31.14 & 75.53 & 8.78 \\
\textbf{\method~(Ours)}     & \textbf{39.08} & \textbf{78.99} & \textbf{8.30} \\
\bottomrule
\end{tabular}
\caption{Ablation study results with different components removed.}
\label{tab:ablation_results}
\end{table}

\subsection{RQ3: Robustness Evaluation}

To evaluate robustness under factual variations, we apply controlled perturbations to case descriptions and examine whether resulting changes in applicable statutes are correctly captured.
This evaluation focuses on \emph{general provisions}, which are more sensitive to some detailed factual changes (e.g., age or mitigating circumstances).
Robustness is quantified using \textit{Change Accuracy}:
\[
\text{Change Accuracy} =
\big|\{\, s \mid s \in S_{\text{pred}} \wedge s \in S_{\text{true}} \,\}\big| / \big| S_{\text{true}} \big|
\]
where \( S_{\text{true}} \) denotes statutes whose applicability changes under perturbation, and \( S_{\text{pred}} \) the statutes predicted after perturbation. Only perturbed statutes are considered.
Table~\ref{tab:robustness_results} reports the robustness performance of all models.
The detailed perturbation settings are provided in the appendix.
\method achieves the highest Change Accuracy (62.56\%),
outperforming all general-purpose and domain-specific baselines.
Domain-specific systems show limited adaptability.
These results indicate that combining agent-based cross-checking with formal logic validation enables more reliable identification of legally relevant changes, yielding more stable statute updates.
Remaining errors mainly stem from upstream fact extraction, suggesting a direction for future improvement.

\begin{table}[ht]
\small
\centering
\begin{tabular}{lr}
\toprule
\textbf{Model} & \textbf{Change Accuracy (\%)} \\
\midrule
LexiLaw                 & 0.00 \\
DISC-LawLLM             & 23.17 \\
GPT o4-mini             & 46.33 \\
GPT-4o                  & 50.67 \\
Claude 4 Sonnet         & 51.50 \\
DeepSeek v3             & 55.93 \\
GPT-5.2                 & 59.63 \\
\textbf{\method~(Ours)} & \textbf{62.56} \\
\bottomrule
\end{tabular}
\caption{Robustness Evaluation on the Perturbation Dataset.}
\label{tab:robustness_results}
\end{table}

\section{Discussion and Limitations}

\paragraph{Error Analysis.}
We observe that the performance for general provisions is consistently lower than that for specific provisions. 
General articles regulate contextual and offender-related factors, such as intent, participation mode, mitigating or aggravating circumstances, rather than explicit criminal acts. 
These elements are often implicitly expressed in case narratives, making them harder to extract and align reliably than act-centered specific statutes.
Due to the inclusion of statute reverse verification, our framework adopts a conservative prediction strategy, favoring legally well-supported statutes while filtering uncertain ones. 
This trade-off is intentional and appropriate for high-stakes legal settings, where avoiding incorrect statute application is more critical than exhaustively enumerating all potentially relevant provisions.

\paragraph{Limitations.}
Despite its promising results, our framework still faces three main challenges. 
First, its formalization quality is constrained by the accuracy of LLM outputs, so errors in identifying actors or conditions can propagate throughout the reasoning pipeline. 
Second, our current evaluation is limited to statutory rules, leaving the extension to case law, open-textured norms, and evolving jurisprudence as future work. 
Third, the system assumes deterministic rule parsing, which prevents it from fully capturing legal provisions that deliberately incorporate normative ambiguity.
In addition, the framework introduces moderate computational overhead due to its multi-stage LLM-based reasoning process.
As summarized in Table~\ref{tab:cost_analysis}, on average, a single case requires multiple LLM calls and over one minute of end-to-end runtime.
Nevertheless, this cost is acceptable given the level of interpretability and formally verified reasoning achieved.

\begin{wraptable}{r}{0.25\textwidth}
\centering
\small
\begin{tabular}{l r}
\hline
\textbf{Metric} & \textbf{Average} \\
\hline
LLM calls per case & 10.33 \\
Runtime (sec) & 107.36 \\
Input tokens & 13{,}049 \\
Output tokens & 4{,}217 \\
Cost per case (USD) & 0.0819 \\
\hline
\end{tabular}
\caption{Average cost/case of \method.}
\label{tab:cost_analysis}
\end{wraptable}


\section{Related Work}

\paragraph{Domain-specific Legal LLMs.}
General-purpose LLMs often mishandle legal terminology and citation style.  
\textsc{ChatLaw} couples a knowledge-graph-enhanced mixture-of-experts backbone with a multi-agent pipeline that mirrors law-firm SOPs, outperforming GPT-4 on LawBench and national bar exams \cite{cui2023chatlaw}.  
\textsc{Lawyer GPT} shows that lightweight domain pre-training plus retrieval boosts statute-matching and consultation accuracy while remaining compute-efficient \cite{yao2024lawyergpt}.  
Gao et al. \cite{gao2024enhancing} further show that careful construction of high-quality synthetic query–candidate pairs can markedly improve legal-case retrieval.
Our work goes further by compiling the extracted norms into \emph{executable} formal reasoning.

\paragraph{Multi-agent Judicial Simulation.}
Treating legal reasoning as collaboration among specialized agents has gained traction.  
\textsc{Agents on the Bench} simulates a collegial bench to improve judgment quality through deliberative voting \cite{jiang2024agentsbench}, whereas \textsc{AgentCourt} evolves adversarial lawyer agents via long-horizon self-play, yielding measurable skill gains \cite{chen2024agentcourt}.  
These systems, however, lack guarantees on logical soundness.  
We retain the agent metaphor and further introduce a \emph{neural–symbolic} pipeline with legal rendering mechanism that integrates statutory interpretation and retrieved precedents..

\paragraph{Neural–Symbolic Methods.}  
Early work such as Logic Tensor Networks \cite{badreddine2020ltn} embeds many-valued fuzzy logic into differentiable architectures, while DeepProbLog \cite{manhaeve2018deepproblog} integrates probabilistic logic programming with neural predicates.  
More recent studies add large-scale LLM components: NS-LCR learns explicit law- and case-level rules for explainable case retrieval \cite{sun2024nslcr}; Logic-LM introduces structured prompting plus theorem-prover feedback to enforce logical soundness \cite{sadowski2025logiclm}; and Kant et al. combine neuro-symbolic reasoning with contract analysis to improve coverage decisions \cite{kant2025robust}.  
These results show that logic grounding enhances transparency and robustness, which we pursue by marrying an agent-driven statute predictor with a symbolic criminal-law reasoner to bridge substantive and formal rationality.

\section{Conclusion}

We presented \method, a solver-centered framework for trustworthy legal AI that bridges substantive and formal rationality in LLM-based legal reasoning.
\method enables legally grounded verdicts that are both interpretable and verifiable.
Experimental results demonstrate that integrating symbolic verification with LLM-based reasoning
improves accuracy, robustness, and explainability over LLM systems.

\newpage
\appendix

\lstset{%
  basicstyle={\footnotesize\ttfamily},
  numbers=left,
  numberstyle=\footnotesize,
  xleftmargin=2em,
  aboveskip=0pt,
  belowskip=0pt,
  showstringspaces=false,
  tabsize=2,
  breaklines=true,
  breakatwhitespace=false,
  columns=fullflexible,
  keepspaces=true
}

\bibliographystyle{named}
\bibliography{ijcai}
\end{document}